\documentclass[conference]{IEEEtran}
\IEEEoverridecommandlockouts
\usepackage{amsmath,amssymb,amsfonts}
\usepackage{algorithmic}
\usepackage{graphicx}
\usepackage{textcomp}
\usepackage{xcolor}
\usepackage{caption}
\usepackage{multirow}
\usepackage{makecell}
\usepackage{cite}
\usepackage{booktabs}
\usepackage{hyperref}
\newcommand{\tabincell}[2]{\begin{tabular}{@{}#1@{}}#2\end{tabular}}
\def\BibTeX{{\rm B\kern-.05em{\sc i\kern-.025em b}\kern-.08em
    T\kern-.1667em\lower.7ex\hbox{E}\kern-.125emX}}
\renewcommand{\footnoterule}{\kern -3pt \hrule width 0.3\linewidth height 0.5pt \kern 2pt}
\begin{document}

\title{Data Analysis and Performance Evaluation of Simulation Deduction Based on LLMs\\
\thanks{*Corresponding author}
}

\author{\IEEEauthorblockN{Shansi Zhang*}
\IEEEauthorblockA{\textit{Artificial Intelligence Institute} \\
\textit{China Electronics Technology Group Corporation}\\
Beijing, 100041, China\\
shansi815@163.com}
\and
\IEEEauthorblockN{Min Li}
\IEEEauthorblockA{\textit{Artificial Intelligence Institute} \\
\textit{China Electronics Technology Group Corporation}\\
Beijing, 100041, China  \\
buptlimin@qq.com}
}

\maketitle

\begin{abstract}
Data analysis and performance evaluation of simulation deduction plays a pivotal role in modern warfare, which enables military personnel to gain invaluable insights into the potential effectiveness of different strategies, tactics, and operational plans. Traditional manual analysis approach is time-consuming and limited by human errors. To enhance efficiency and accuracy, large language models (LLMs) with strong analytical and inferencing capabilities can be employed. However, high-quality analysis reports with well-structured formatting cannot be obtained through a single instruction input to the LLM. To tackle this issue, we propose a method that first decomposes the complex task into several sub-tasks and designs effective system prompts and user prompts for each sub-task. Multi-round interactions with the LLM incorporating self-check and reflection are then conducted to enable structured data extraction as well as multi-step analysis and evaluation. Furthermore, custom tools are defined and invoked to generate figures and compute metrics. We also design multiple report templates, each tailored to a specific application and input data type, ensuring their adaptability across a variety of scenarios. Extensive evaluation results demonstrate that the reports generated by our method exhibit higher quality, therefore obtaining higher scores than the baseline method. 
\end{abstract}


\section{Introduction}
Analyzing and evaluating military simulation deduction data is crucial for understanding the effectiveness of strategies, identifying potential weaknesses, and making informed decisions~\cite{Shi2021}. It enables military personnel to assess the impact of various factors on mission outcomes and improve their approaches. This process is essential for optimizing resource allocation and preparing for real-world scenarios. 

Traditional manual analysis, however, is limited by its time-consuming nature, susceptibility to human errors, and inability to process vast amounts of data quickly. This is where large language models (LLMs) come in. LLMs can analyze data at an unprecedented scale and speed, providing more accurate and comprehensive insights. They can identify patterns and trends that might be overlooked by human analysts and offer recommendations based on a deeper understanding of the data~\cite{Hong2024b}. Incorporating LLMs into military simulation analysis is not just an advantage; it's a necessity for staying ahead in an increasingly complex and data-driven environment. Using advanced LLM-based analytics not only enhances the quality of insights but also the speed and efficiency with which they are delivered, making it imperative for modern military organizations to adopt such technologies~\cite{Hua2024}.

Although LLMs have the capability to provide accurate and efficient analysis, high-quality and well-formatted reports cannot be generated through a single instruction input alone. To address this limitation, we propose a method that involves task decomposition and sub-task prompt design, and performs multi-round analysis and evaluation using the LLM with self-check and reflection mechanism. Additionally, specialized tools are integrated to produce visual charts and calculate metrics, thereby generating reports with complete analysis and well-illustrated figures. We also develop multiple report templates customized for different applications and input data types, which can be flexibly adapted to various scenarios. Our main contributions are summarized as follows:
\begin{itemize}
\item We decompose complex report generation tasks into several sub-tasks, and design effective system prompts and user prompts based on these sub-tasks for various types of report generation, which enable LLMs to clarify their roles, understand corresponding rules, and generate comprehensive analysis contents.
\item We design effective pipelines to generate multiple categories of reports through multi-round interactions for structured data extraction and multi-step analysis and evaluation. Plotting and calculation tools are invoked to achieve visualization and metric calculation, respectively.
\item We conduct extensive evaluations, which demonstrate that the reports generated by our method exhibit higher quality and better formatting, thus achieving significantly higher scores compared to the baseline method.
\end{itemize}

\section{Related Work}
\subsection{Applications of LLMs in the military domain}
LLMs hold significant potential in the military domain. Caballero et al.~\cite{Caballero2025} proposed an examination of the transformative potential and strategic implications of integrating LLMs into national security contexts, highlighting their ability to enhance information processing, decision-making, and operational efficiency. Liu et al.~\cite{Liu2024} proposed MilChat, a military information assistance system based on an LLM, designed to provide personnel with accurate and standardized information about enemy equipment in offline environments. The system was built upon the open-source Qwen-7B-Chat model~\cite{Qwen2024}, which was fine-tuned using LoRA~\cite{Hu2022} and military-specific datasets to improve performance. Hong et al.~\cite{Hong2024a} developed an LLM-based prompt framework for selecting replanning groups in combat systems to effectively respond to unexpected situations during real combat. By integrating comprehensive information about agents, tasks, and unforeseen events—such as agent status, task importance and urgency, and situational details—the system leverages LLMs to analyze and select the most suitable agents and tasks for replanning. Goecks et al.~\cite{Goecks2024} proposed COA-GPT, a novel algorithm that leverages LLMs to rapidly generate and refine Courses of Action (COAs) in military operations. By integrating military doctrine and domain expertise through in-context learning, the system allows commanders to input mission information in text or image format and receive strategically aligned COAs within seconds. Zhang et al.~\cite{Zhang2023} proposed a simulation scenario generation method based on LLMs and knowledge graphs. By integrating prompt engineering with domain-specific knowledge, the approach utilizes LLMs to generate structured simulation scenarios, with the knowledge graph serving as a guiding framework. Oijen et al.~\cite{Oijen2025} developed a conceptual framework that utilizes LLMs to support human-AI collaboration in military simulation-based training scenario development, aiming to improve efficiency and effectiveness while maintaining instructional integrity.

\subsection{LLMs for data analysis}
Due to their powerful analytical and inferential capabilities, LLMs can efficiently assist people in a wide variety of data analysis tasks. Liu et al.~\cite{Liu2023} introduced JarviX, a sophisticated data analytics framework designed to leverage LLMs to provide automated guidance and perform high-precision data analysis on tabular datasets. Guo et al.~\cite{Guo2024} proposed two probes, an open-ended high agency prototype and a structured low agency prototype, to explore the application of LLMs in domain-specific data analysis tools from the dimensions of interaction and user agency. Pérez et al.~\cite{Perez2025} developed a framework that leverages LLMs to automatically generate textual insights from multi-table databases. Their approach includes a Hypothesis Generator for formulating domain-relevant questions, a Query Agent for generating and executing SQL queries to answer those questions, and a summarization module for verbalizing the results into concise, insightful text. Zhang et al.~\cite{Zhang2025} proposed an LLM-driven workflow for geoscience data analysis that addresses the challenges of adapting LLMs to domain-specific requirements without the need for fine-tuning or extensive human alignment. Almheiri et al.~\cite{Almheiri2024} proposed a framework for conducting data analytics using LLMs, which is based on four fundamental constructs: meta-specifications, specifications, instructions, and prompting patterns. This framework provides a structured approach for data engineers, analysts, and domain experts to interact with LLMs. Wang et al.~\cite{Wang2025} developed an intelligent financial data analysis system that integrates LLMs with Retrieval-Augmented Generation (RAG)~\cite{Lewis2020} technology to overcome the limitations of traditional financial data analysis methods. The system includes a financial data preprocessing module, a vector-based storage and retrieval system, and a RAG-enhanced query processing module.

\section{Proposed Method}

\subsection{Overview}

\begin{figure*}[htbp]
\centerline
{\includegraphics[width=1\textwidth]{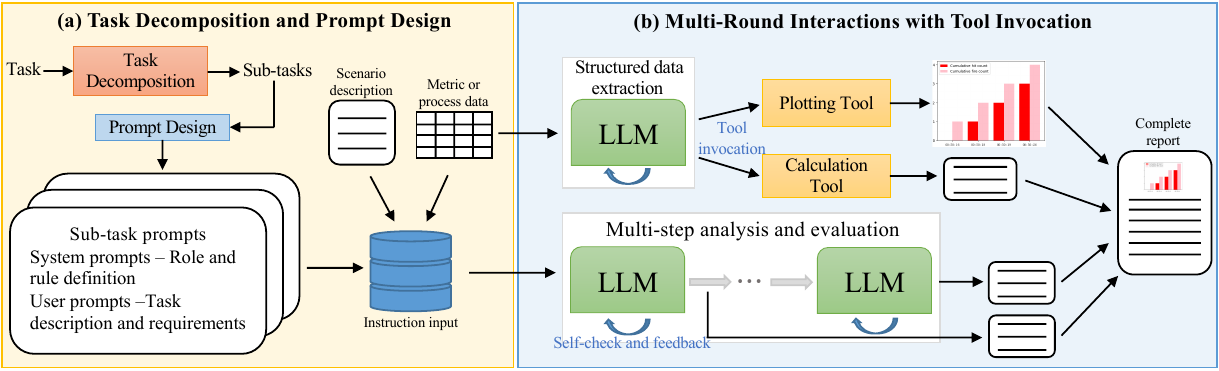}}
\caption{The overall pipeline of report generation.}
\label{fig:pipeline}
\end{figure*}

Our objective is to leverage LLMs to generate diverse types of analysis and evaluation reports, utilizing metric or process data derived from simulation deduction. The overall pipeline of our proposed method is illustrated in Figure~\ref{fig:pipeline}. It consists of two key components: (1) \textit{Task decomposition and prompt design}, where the report generation task is decomposed into several sub-tasks and specialized prompts are crafted for each; (2) \textit{Multi-round interactions with tool invocation}, which involves multi-step analysis and evaluation using the LLM while supporting tool invocation for plotting figures and calculate metrics. Together, these two components enable the production of complete and aesthetic reports. In what follows, we will introduce the above techniques in detail. 

\subsection{Task Decomposition and Prompt Design}
To accommodate diverse applications and data inputs, we employ LLMs to design efficient workflows for generating five distinct report categories: (1) Report on effectiveness evaluation and optimization suggestions for red-blue confrontation operations (Report A); (2) Report on comparative analysis and configuration suggestions of multi-factor experiments (Report B); (3) Report on event-driven analysis of the operational process and capability assessment (Report C); (4) Report on comprehensive analysis of operational processes in multiple trials of a certain case (Report D); (5) Report on comprehensive comparative analysis and configuration suggestions of multi-factor experiments (Report E). The data inputs for each type of report are listed in Table~\ref{tab:report types}.

\begin{table}[htbp]
\caption{Data inputs for different types of reports}
\begin{center}
\begin{tabular}{c|c}
\toprule
Report type & Data input\\
\midrule
Report A & \tabincell{c}{Scenario description, metric data \\of the red and blue sides in a single trial}\\
\midrule
Report B & \tabincell{c}{Scenario and factor description, \\ metric data of multi-factor experiments}\\
\midrule
Report C & \tabincell{c}{Scenario description, multiple process \\data files in a single trial}\\
\midrule
Report D & \tabincell{c}{Scenario description, process data files \\of multiple trials under a certain case}\\
\midrule
Report E & \tabincell{c}{Scenario and factor description, \\summaries of process data analysis in all cases}\\
\bottomrule
\end{tabular}
\end{center}
\label{tab:report types}
\end{table}

To generate comprehensive reports with in-depth analysis and rich visualizations, we employ a task decomposition strategy that breaks down the report generation task into several sub-tasks, including structured data extraction, multi-step analysis and evaluation, and metric calculation. For each sub-task, we design specialized prompts to guide the corresponding content generation. This approach can effectively overcome the context token limitation of a single invocation to the LLM and alleviate its generation burden, thereby facilitating the production of higher-quality and more reliable reports.

The prompts consist of system prompts and user prompts. System prompts contain the core settings of the model (such as role positioning, capability boundaries, format requirements, etc.), which are usually pre-implanted by developers and mostly used for model initialization and defining long-term rules. For example, the system prompts designed for comparative analysis experiments are as follows: \textit{You are an expert in result analysis and evaluation of simulation deduction, responsible for analyzing and evaluating trial results. You need to accurately identify the changes in metrics under different trial settings and provide reasonable cause analysis. Please carefully check before outputting the analysis and evaluation results to avoid obvious errors in numerical change judgment.} These prompts define the rules and responsibilities of the LLM, while also requiring it to perform self-check and reflection. 

User prompts are instructions given to the model for specific tasks, influencing the contents of single-time outputs. They are composed by users based on immediate needs, focusing on specific issues with flexible adjustments to details. For example, the user prompts designed for comparative analysis experiments are as follows: \textit{Suppose multiple sets of comparative trials have been carried out, and the descriptions of the scenario and experimental factor changes are as follows: \{\}. The metrics of the red and blue sides under multiple sets of trials are as follows: \{\}. Please conduct a comparative analysis of the metrics of the red and blue sides under different trial schemes, first analyze the influence degree of each change factor on the metrics and the underlying reasons, then give the comprehensive and conclusive analysis results, and finally provide reasonable configuration suggestions for the red and blue sides.} The partial user prompts designed for comprehensive analysis experiments of operational processes are as follows: \textit{Based on the operational summaries from multiple trials, please generate a table containing the hit rate metrics and operational outcome summaries for both the red and blue sides in each trial. Then, based on the table, summarize the performance of both sides under this experimental setup, focusing on their average hit rates and the number of successfully completed missions.} These prompts clearly specify the concrete task requirements and expected output content. 

\subsection{Multi-Round Interaction with Tool Invocation}
To generate complex reports, instead of invoking the LLM only once to output all contents, we adopt a multi-round interaction approach, where each interaction aims to complete a specific sub-task. The LLM performs self-check and reflection through prompt instructions to ensure the correctness of the output content. The outputs of all the interactions are subsequently synthesized into a final report. Notably, since LLMs lack native capabilities for data visualization or metric computation, we implement external tool integration. This involves: (1) developing specialized functions for plotting and calculations, (2) binding these tools to the LLM, and (3) designing prompts that enable autonomous tool invocation – thereby achieving integrated visual analytics and quantitative assessment.

We design five types of report templates for evaluating simulation deduction. Depending on the complexity of each report, a varying number of interactions are required to generate high-quality outputs. For instance, to generate report B, the LLM first extracts the data specified by the user for visualization and returns it in a structured format in the first round, after which a plotting tool is used to create corresponding figures. In the second round, the LLM performs analysis and evaluation according to the user's instruction to generate concrete contents, including experimental factor influences, comprehensive analysis results, and configuration suggestions.

To generate report C, the LLM first extracts structured data for visualization in the initial round. In the second round, it reconstructs the complete operational process and provides a summary based on the scenario description and various types of process information data. In the third round, the LLM invokes the associated tool to calculate the hit rates of the red and blue sides, and evaluates the operational capabilities of both sides according to the computed results.

To generate report D, the LLM is invoked multiple times to reconstruct the operational process, summarize the outcomes, and calculate metrics for each trial under a certain experimental setting. Figures are plotted to visually represent the metrics across trials. Subsequently, the LLM creates a table that includes the metrics of the red and blue sides and a summary of the operational results for each trial. Finally, the LLM provides an overall assessment of the performance of both sides under this experimental setting, including average metrics and the number of times the goal is achieved. 

To generate report E, the LLM first extracts structured metric data from the summary reports of each case for visualization. Then, the LLM conducts a comprehensive analysis to explore the impact of each experimental factor and provide reasonable configuration recommendations. 

\section{Experiments}

\subsection{Dataset}
The data used in our experiments were generated through simulation deduction. We first established a basic scenario by configuring the natural environment, the equipment of the red and blue sides, as well as unit behaviors. Based on this basic scenario, we adjusted specific factor levels to conduct comparative experiments. There were five cases in total with different experimental settings, and ten trials were conducted for each case. During each trial, multiple process data files that recorded the changes of various states over time were saved. Additionally, the evaluation system integrated into the simulation platform can parse these process data files and calculate key performance metrics for both the red and blue sides.

\subsection{Implementation Details}
We designed customized templates for each type of report and implemented them using the LangChain framework. Each report type corresponds to a configuration file, which primarily includes file paths, user-specified data to be visualized, and additional prompts provided by the user. These report templates are adaptable to various scenarios by simply adjusting the parameters in the configuration files. We deployed multiple open-source LLMs on the server, including DeepSeek-R1-70B~\cite{Guo2025}, Qwen3-32B~\cite{Yang2025}, and Qwen3-235B-A22B~\cite{Yang2025}. All selected LLMs are reasoning models, as they excel more in data analysis, cause exploration, and suggestion generation. NVIDIA RTX A6000 GPUs were employed to run these LLMs. 

\subsection{Results}
To quantitatively evaluate the quality of generated reports, we defined a set of scoring criteria. During the scoring process, four key aspects were considered: data analysis accuracy, content completeness, practicality, and layout aesthetics. The weights assigned to these aspects are 0.3, 0.2, 0.3, and 0.2, respectively, reflecting their relative importance, with each aspect scored on a scale of 1 to 10. Consequently, the overall score is normalized to a maximum of 10 points. We invited three human evaluators and employed three LLM products (DeepSeek-R1-0528, Wenxin-4.5-Turbo, and Doubao-1.5) to score the reports based on the same criteria. The final scores were determined by calculating the average scores from both the human evaluators and the LLMs. We compared our method against a baseline method, which used the same models as ours but did not incorporate task decomposition, multi-step analysis and evaluation, and tool invocation. 

The quantitative results are listed in Table~\ref{tab:results}. It can be seen that the reports generated by our method achieve higher scores than those produced by the baseline method, and the performance advantage becomes more pronounced as the complexity of the reports increases. For instance, reports B, C, D, and E generated using our method, which involves multiple interactions with the LLM, attain significantly higher scores than those from the baseline method, regardless of whether they are evaluated by human users or LLMs. Additionally, the reports generated by  Qwen3-235B-A22B typically achieve higher scores than those generated by DeepSeek-R1-70B and Qwen3-32B, which suggests that more advanced LLMs contribute to the generation of higher-quality reports. The time required to generate reports varies depending on their complexity, typically ranging from approximately 1 to 5 minutes.

\begin{table*}[htbp]
\caption{Evaluation results of different types of reports}
\begin{center}
\begin{tabular}{cc|cc|cc|cc|cc|cc}
\toprule
\multirow{2}{*}{Methods} & \multirow{2}{*}{Models} & \multicolumn{2}{c|}{Report A} & \multicolumn{2}{c|}{Report B} & \multicolumn{2}{c|}{Report C} & \multicolumn{2}{c|}{Report D} & \multicolumn{2}{c}{Report E}\\
\cline{3-12}
& & Users & LLMs & Users & LLMs & Users & LLMs & Users & LLMs & Users & LLMs\\
\midrule
\multirow{3}{*}{Ours} & DeepSeek-R1-70B & 8.167 & 7.933 & 7.267 & 7.700 & 8.333 & 7.633 & 7.467 & 8.100 & 7.500 & 7.600 \\
& Qwen3-32B & 8.067 & 7.700 & 7.233 & 7.567 & 8.200 & 7.633 & 7.367  & 7.900 & 7.167 & 7.667 \\
& Qwen3-235B-A22B & 8.967 & 8.433 & 8.167 & 7.733 & 8.333 & 7.867 & 7.833 & 8.067 & 7.867 & 8.867 \\
\midrule
\multirow{3}{*}{Baseline} & DeepSeek-R1-70B & 7.800 & 7.533 & 5.733 & 6.233 & 6.300 & 5.933 & 5.433 & 5.700 & 5.533 & 5.633 \\
& Qwen3-32B & 7.700 & 7.533 & 5.767 & 6.133 & 6.167 & 6.033 & 5.533 & 5.867 & 5.700 & 5.867 \\
& Qwen3-235B-A22B & 8.267 & 8.233 & 6.633 & 6.600 & 6.367 & 6.067 & 5.967 & 6.033 & 6.067 & 6.267\\
\bottomrule
\end{tabular}
\end{center}
\label{tab:results}
\end{table*}

\begin{figure}[htbp]
\centerline
{\includegraphics[width=0.5\textwidth]{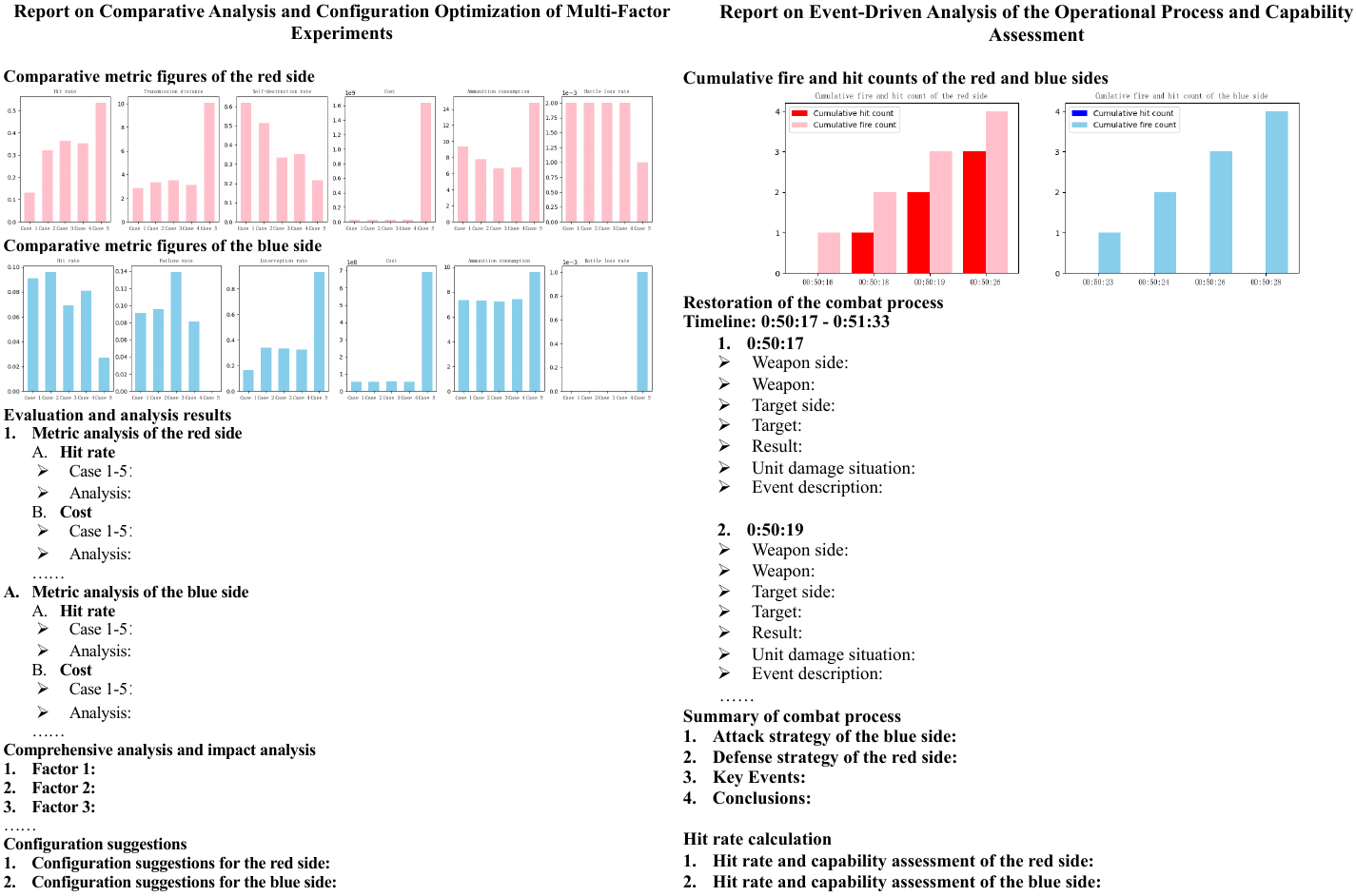}}
\caption{Templates of generated reports.}
\label{fig:template}
\end{figure}

Figure~\ref{fig:template} presents the templates for the generated reports B and C by our method. As shown, the reports include visual charts that intuitively illustrate the performance of the red and blue sides, along with comprehensive analysis and evaluation of the metrics or operational processes.

\section{Conclusions}
In this paper, we propose a method that leverages LLMs to achieve data analysis and performance evaluation of simulation deduction automatically. To produce high-quality, well-formatted analysis reports, we first break down the complex task into several sub-tasks, designing system prompts and user prompts for each. Then, multi-round interactions with the LLM incorporating self-check and reflection are carried out to facilitate structured data extraction alongside multi-step analysis and evaluation. Additionally, we define and employ custom tools to generate diagrams and calculate metrics. Moreover, we create a variety of report templates tailored to different applications and data inputs, ensuring versatility across diverse scenarios. Comprehensive evaluation results show that our method generates reports of superior quality, achieving higher scores compared to the baseline approach.


\bibliographystyle{IEEEtran}
\bibliography{references}

\end{document}